# Data-Driven Transfer Learning Framework for Estimating Turning Movement Counts

**Xiaobo Ma, Ph.D.**
Pima Association of Governments
1 E Broadway Blvd., Ste. 401
Tucson, AZ 85701
Email: xiaoboma@arizona.edu

**Hyunsoo Noh, Ph.D.**
Pima Association of Governments
1 E Broadway Blvd., Ste. 401
Tucson, AZ 85701
Email: HNoh@pagregion.com

**Ryan Hatch**
Pima Association of Governments
1 E Broadway Blvd., Ste. 401
Tucson, AZ 85701
Email: RHatch@pagregion.com

**James Tokishi**
Pima Association of Governments
1 E Broadway Blvd., Ste. 401
Tucson, AZ 85701
Email: JTokishi@pagregion.com

**Zepu Wang**
Civil & Environmental Engineering
University of Washington
3760 E. Stevens Way NE
Seattle, WA 98195
Email: zepu@uw.edu

**ABSTRACT**

Urban transportation networks are vital for the efficient movement of people and goods, necessitating effective traffic management and planning. An integral part of traffic management is understanding the turning movement counts (TMCs) at intersections, Accurate TMCs at intersections are crucial for traffic signal control, congestion mitigation, and road safety. In general, TMCs are obtained using physical sensors installed at intersections, but this approach can be cost-prohibitive and technically challenging, especially for cities with extensive road networks. Recent advancements in machine learning and data-driven approaches have offered promising alternatives for estimating TMCs. Traffic patterns can vary significantly across different intersections due to factors such as road geometry, traffic signal settings, and local driver behaviors. This domain discrepancy limits the generalizability and accuracy of machine learning models when applied to new or unseen intersections. In response to these limitations, this research proposes a novel framework leveraging transfer learning (TL) to estimate TMCs at intersections by using traffic controller event-based data, road infrastructure data, and point-of-interest (POI) data. Evaluated on 30 intersections in Tucson, Arizona, the performance of the proposed TL model was compared with eight state-of-the-art regression models and achieved the lowest values in terms of Mean Absolute Error and Root Mean Square Error.

# 1. INTRODUCTION

Urban transportation networks are the lifelines of modern cities, supporting the movement of people and goods. Efficient traffic management and planning are paramount for ensuring smooth and safe travel within these networks(Cottam et al., 2024; X. Ma, Karimpour, et al., 2023b). The accurate estimation of Turning Movement Counts (TMCs) at signalized intersections is a critical component in urban traffic management and planning. TMCs provide essential data for traffic signal control, congestion mitigation, and road safety enhancement(X. Ma, Karimpour, et al., 2023a; Wu et al., 2021, 2022). Traditional methods of collecting TMC data often involve manual counting, which can be costly, time-consuming, and susceptible to human error. As cities grow and traffic patterns become more complex, there is a pressing need for more efficient, reliable, and scalable methods to estimate TMCs.

Loop detectors, which are embedded in the pavement to detect the presence of vehicles, are commonly used for this purpose due to their reliability and relatively low cost(X. Ma et al., 2020; X. Ma, Karimpour, et al., 2023c; Wu et al., 2019). Ghods and Fu (2014) proposed a method to estimate TMCs at signalized intersections based on entry/exit traffic volumes collected from loop detectors and signal phase information (Ghods & Fu, 2014). Gholami and Tian (2016) used a network equilibrium approach to process loop detector data for estimating TMCs at shared lanes(Gholami & Tian, 2016). In another study, loop detector data combined with an adaptive neural fuzzy inference system and genetic programming were applied to estimate TMCs (Gholami & Tian, 2017) . A recent study employed data collected from conventional long-loop detectors for left-turn movement count estimation based on machine-learning models (Biswas et al., 2022). However, implementing these methods on a region-wide scale poses significant challenges. One of the primary obstacles is that the majority of intersections are not equipped with loop detectors. Without these detectors, accurately estimating TMCs at numerous intersections becomes an unfeasible endeavor.

The application of video cameras and computer vision technologies for estimating TMCs at intersections has gained substantial interest in recent years(Lin et al., 2024). Video cameras offer a versatile and non-intrusive means of capturing real-time traffic data, providing a comprehensive view of vehicular movements(Hu et al., 2023; Sun et al., 2024). Coupled with advancements in computer vision, these systems can automatically detect and track vehicles, classify them by type, and accurately count turning movements. Studies have demonstrated that computer vision algorithms, including convolutional neural networks (CNNs) and deep learning techniques, can achieve high levels of accuracy in traffic flow analysis. For instance, research by Shirazi and Morris (2016) highlighted the efficacy of using a vision-based vehicle tracking system to analyze video footage and estimate TMCs (Shirazi & Morris, 2016). Bélisle et al., (2017) introduced an innovative method for automatically counting vehicle turning movements using video tracking. It builds upon previous research that focused on optimizing parameters for extracting road user trajectories and automating the clustering of these trajectories (Bélisle et al., 2017). Adl et al., (2024) proposed an innovative framework for vehicle counting at intersections using a fisheye camera system. Their experiments on real-world datasets yielded promising results, underscoring the framework's potential for applications in intelligent traffic control and urban planning (Adl et al., 2024). Despite the advantages brought by video cameras and computer vision technologies, several challenges persist, such as the necessity for high-quality video footage, the development of robust algorithms capable of performing reliably under diverse weather conditions and lighting, and the significant computational resources required for real-time processing. Additionally, the high costs associated with advanced video and sensor technologies present significant financial barriers, preventing broad deployment despite their potential benefits. Thus, the widespread adoption of video cameras for TMC estimation is currently not feasible for many transportation agencies, necessitating the exploration of more cost-effective and scalable alternatives.

Recent advancements in machine learning and data-driven approaches have offered promising alternatives for estimating TMCs(X. Ma, 2022; Z. Zhang et al., 2024). Various studies have explored the use of supervised learning techniques, such as regression models and neural networks, to predict vehicle movements based on historical traffic data and real-time inputs(Wang et al., 2024; Yang et al., 2024). For example, Ghanim and Shaaban (2019) utilized artificial neural networks (ANN) to analyze the relationship between approach volumes and the corresponding turning movements, leveraging large datasets to train their algorithms (Ghanim & Shaaban, 2019). Another study involves the use of support vector regression (SVR), random forest (RF), and ANN to improve estimation accuracy and robustness (Biswas et al., 2022). Xu et al., (2023) demonstrated that the well-calibrated ANN could effectively estimate TMCs by integrating traffic controller event-based data (Xu et al., 2023). Machine learning methods are scalable and can be deployed across multiple intersections, offering a cost-effective solution compared to traditional methods such as manual counting or loop detectors, as well as emerging technologies like video cameras and computer vision(Zhao et al., 2023, 2024). Despite these advancements, conventional machine learning models often assume that the data distributions of training and testing datasets are identical, which is seldom the case in real-world scenarios. Traffic patterns can vary significantly across different intersections due to factors such as road geometry, traffic signal settings, and local driver behaviors. This domain discrepancy limits the generalizability and accuracy of traditional models when applied to new or unseen intersections(H. Ma, Zeng, et al., 2023).

To overcome the above-mentioned limitations, this research proposes a data-driven transfer learning (TL) framework for estimating TMCs. The concept of TL has gained significant traction in recent years, primarily in the fields of machine learning and artificial intelligence. TL is an advanced machine learning technique that leverages knowledge from related tasks or domains to improve the performance of a model on a target task(X. Ma, Cottam, et al., 2023). By relaxing the assumption that source and target data distributions must be identical, TL can adapt pre-trained models to new domains with limited data, thus enhancing prediction accuracy and generalizability (X. Ma et al., 2024). This approach is particularly valuable in scenarios where collecting new data is expensive or impractical. In the context of traffic management, TL can harness data from well-instrumented intersections (source domain) to infer traffic patterns at intersections lacking physical sensors (target domain).

The proposed framework accurately estimates TMCs at intersections by using traffic controller event-based data, road infrastructure data, and point-of-interest (POI) data. In addition, the proposed framework develops scene-specific models by employing a TL model. Because the TL model relaxes the assumption that the underlying data distributions of the source and target domains must be the same, the proposed framework can guarantee high-accuracy estimation of TMCs for intersections with different traffic patterns, distributions, and characteristics. The performance of the proposed framework is evaluated using 30 intersections in Tucson, Arizona.

This research makes several key contributions to the field of traffic management and TMC estimation:

- **Introduction of a TL framework for TMC estimation:** This study pioneers the application of TL in the context of TMC estimation, demonstrating its potential to improve accuracy and efficiency.
- **Comprehensive evaluation against traditional models:** The proposed TL framework will be rigorously compared with various regression models, highlighting its advantages and addressing potential challenges.
- **Development of scene-specific models:** By tailoring models to specific intersections, the framework provides more precise and context-aware TMC estimates, facilitating better traffic management decisions.
- **Scalability and adaptability:** This research presents a scalable solution that can continuously adapt to evolving traffic conditions. Urban traffic systems are dynamic, with constantly changing traffic conditions. TL frameworks can be more easily updated and adapted to new data, providing a scalable solution for continuous traffic monitoring and management. This research primarily utilizes traffic controller event-based data, which offers the advantage of region-wide coverage and low cost. This is because most local transportation agencies have already configured traffic detectors for actuated signal control.
- **Promote data efficiency:** Collecting extensive TMC data for every intersection is impractical. TL allows models to learn from limited data by leveraging related information from other intersections, reducing the dependency on large datasets.
- **Guarantee resource optimization:** Implementing physical sensors at every intersection is cost-prohibitive. A TL-based approach offers a cost-effective alternative by maximizing the utility of existing data and reducing the need for additional infrastructure.

In conclusion, the proposed data-driven TL framework represents a significant advancement in the estimation of TMCs. By addressing the limitations of conventional models and leveraging the strengths of TL, this research aims to provide a robust, efficient, and scalable solution for modern urban traffic management.

The remainder of this paper is organized as follows: preliminaries and problem definition are described in the second section. The third section introduces the proposed framework and methodologies. In the fourth section, a case study is presented to examine the transferability of the proposed framework. Lastly, section five provides the conclusion and future work.

## 2. PROBLEM DEFINITION AND PRELIMINARIES

The notations, traffic variable definitions, and data encoding are presented first. Subsequently, the problem addressed in this study is formulated.

### 2.1. Notations

Table 1 lists all the variables and their definitions that were used for the model development.

**Table 1 Notations and Variables**

| Categories | Variables | Definition |
|---|---|---|
| Event information for through movement | $o_{TM}^{a,i,t}$ | Detector occupancy time for through movement of approach $a$ of intersection $i$ at time interval $t$ |
| | $d_{TM}^{a,i,t}$ | Detector trigger counts for through movement of approach $a$ of intersection $i$ at time interval $t$ |
| | $g_{TM}^{a,i,t}$ | Green time duration for through movement of approach $a$ of intersection $i$ at time interval $t$ |
| | $c_{TM}^{a,i,t}$ | Cycle counts for through movement of approach $a$ of intersection $i$ at time interval $t$ |
| | $m_{TM}^{a,i,t}$ | Average of time differences between each pair of consecutive detections for through movement of approach $a$ of intersection $i$ at time interval $t$ |
| | $s_{TM}^{a,i,t}$ | Standard deviation of time differences between each pair of consecutive detections for through movement of approach $a$ of intersection $i$ at the time interval $t$ |
| Event information for left-turn movement | $o_{LM}^{a,i,t}$ | Detector occupancy time for left-turn movement of approach $a$ of intersection $i$ at time interval $t$ |
| | $d_{LM}^{a,i,t}$ | Detector trigger counts for left-turn movement of approach $a$ of intersection $i$ at time interval $t$ |
| | $g_{LM}^{a,i,t}$ | Green time duration for left-turn movement of approach $a$ of intersection $i$ at time interval $t$ |
| | $c_{LM}^{a,i,t}$ | Cycle counts for left-turn movement of approach $a$ of intersection $i$ at time interval $t$ |
| | $m_{LM}^{a,i,t}$ | Average of time differences between each pair of consecutive detections for left-turn movement of approach $a$ of intersection $i$ at time interval $t$ |
| | $s_{LM}^{a,i,t}$ | Standard deviation of time differences between each pair of consecutive detections for left-turn movement of approach $a$ of intersection $i$ at the time interval $t$ |
| | $p_{LM}^{a,i,t}$ | Permissive green time for left-turn movement of approach $a$ of intersection $i$ at time interval $t$ |
| Intersection layout information | $l_{SL}^{a,i}$ | Number of shared left turn lanes of approach $a$ of intersection $i$ |
| | $l_{EL}^{a,i}$ | Number of exclusive left turn lanes of approach $a$ of intersection $i$ |
| | $l_{TL}^{a,i}$ | Number of through lanes of approach $a$ of intersection $i$ |
| | $l_{ER}^{a,i}$ | Number of exclusive right turn lanes of approach $a$ of intersection $i$ |
| | $l_{SR}^{a,i}$ | Number of shared right turn lanes of approach $a$ of intersection $i$ |
| | $e_{POIE}^{i}$ | Number of employees of all POI within 400 m of intersection $i$ |
| POI information | $e_{POIC}^{i}$ | POI categories count within 400 m of intersection $i$ |
| Turning movement counts | $v_{LM}^{a,i,t}$ | Traffic counts for left-turn movement of approach $a$ of intersection $i$ at the time interval $t$ |
| | $v_{TM}^{a,i,t}$ | Traffic counts for through movement of approach $a$ of intersection $i$ at the time interval $t$ |
| | $v_{RM}^{a,i,t}$ | Traffic counts for right-turn movement of approach $a$ of intersection $i$ at the time interval $t$ |

## 2.2. Data encoding

To make road type, left-turn type, and datetime accessible by the model, data are encoded as follows:

**(1) Road type**

The major road is encoded as 1 and minor road is encoded as 2. Road type information for approach $a$ of intersection $i$ is denoted as $r^{a,i}$.

**(2) Left-turn type**

Considering there are different types of left-turn phases, permissive-only left-turn is encoded as 1, protected-permissive left-turn is encoded as 2, and protected-only left-turn is encoded as 3. Left-turn type information for approach $a$ of intersection $i$ is denoted as $l^{a,i}$.

**(3) Minute-of-hour and hour-of-day**

Each hour has four 15-minute intervals, therefore, the minute-of-hour (MOH) takes values from 1 to 4 to represent each 15-minute interval. Hour-of-day (HOD) is represented by subsequent numbers, starting from 0 to 23 (0 representing midnight, and 23 representing 11 pm). MOH information and HOD information for the time interval $t$ are denoted as $h_{MOH}^{t}$ and $h_{HOD}^{t}$.

## 2.3. Problem formulation

The purpose of this study is to build a TL framework to estimate TMCs for intersections. This goal can be achieved through training models on a known intersection (henceforth referred to as "source domain") and transferring the well-trained models to estimate TMCs for a new intersection (henceforth referred to as "target domain"). As shown in Eq. (1), Eq. (2), and Eq. (3), by associating traffic variables for a known intersection, the learned function $F_1(\cdot)$, $F_2(\cdot)$, and $F_3(\cdot)$, can be used for left-turn, through, and right-turn movement counts estimation on a new intersection.

$$F_1\left(\begin{bmatrix} o_{TM}^{a,i,t}, d_{TM}^{a,i,t}, g_{TM}^{a,i,t}, c_{TM}^{a,i,t}, m_{TM}^{a,i,t}, s_{TM}^{a,i,t}, o_{LM}^{a,i,t}, d_{LM}^{a,i,t}, g_{LM}^{a,i,t}, c_{LM}^{a,i,t}, m_{LM}^{a,i,t}, \\ s_{LM}^{a,i,t}, p_{LM}^{a,i,t}, l_{SL}^{a,i}, l_{EL}^{a,i}, l_{TL}^{a,i}, l_{ER}^{a,i}, l_{SR}^{a,i}, e_{POIE}^{i}, e_{POIC}^{i}, r^{a,i}, l^{a,i}, h_{MOH}^{t}, h_{HOD}^{t} \end{bmatrix}\right) = [v_{LM}^{a,i,t}] \quad (1)$$

$$F_2\left(\begin{bmatrix} o_{TM}^{a,i,t}, d_{TM}^{a,i,t}, g_{TM}^{a,i,t}, c_{TM}^{a,i,t}, m_{TM}^{a,i,t}, s_{TM}^{a,i,t}, o_{LM}^{a,i,t}, d_{LM}^{a,i,t}, g_{LM}^{a,i,t}, c_{LM}^{a,i,t}, m_{LM}^{a,i,t}, \\ s_{LM}^{a,i,t}, p_{LM}^{a,i,t}, l_{SL}^{a,i}, l_{EL}^{a,i}, l_{TL}^{a,i}, l_{ER}^{a,i}, l_{SR}^{a,i}, e_{POIE}^{i}, e_{POIC}^{i}, r^{a,i}, l^{a,i}, h_{MOH}^{t}, h_{HOD}^{t} \end{bmatrix}\right) = [v_{TM}^{a,i,t}] \quad (2)$$

$$F_3\left(\begin{bmatrix} o_{TM}^{a,i,t}, d_{TM}^{a,i,t}, g_{TM}^{a,i,t}, c_{TM}^{a,i,t}, m_{TM}^{a,i,t}, s_{TM}^{a,i,t}, o_{LM}^{a,i,t}, d_{LM}^{a,i,t}, g_{LM}^{a,i,t}, c_{LM}^{a,i,t}, m_{LM}^{a,i,t}, \\ s_{LM}^{a,i,t}, p_{LM}^{a,i,t}, l_{SL}^{a,i}, l_{EL}^{a,i}, l_{TL}^{a,i}, l_{ER}^{a,i}, l_{SR}^{a,i}, e_{POIE}^{i}, e_{POIC}^{i}, r^{a,i}, l^{a,i}, h_{MOH}^{t}, h_{HOD}^{t} \end{bmatrix}\right) = [v_{RM}^{a,i,t}] \quad (3)$$

Given source domain data $D_S = \{(x_{S_1}, y_{S_1}), \dots, (x_{S_n}, y_{S_n})\}$, where $x_{S_i} \in \mathcal{X}_S$ is the data instance and $y_{S_i} \in \mathcal{Y}_S$ is the corresponding label. $F_1(\cdot)$, $F_2(\cdot)$, and $F_3(\cdot)$ can be generalized as $F_4(\cdot)$.

$$F_4([\mathcal{X}_S]) = [\mathcal{Y}_S] \quad (4)$$

Let target domain data $D_T = \{(x_{T_1}, y_{T_1}), \dots, (x_{T_k}, y_{T_k})\}$, where $x_{T_i} \in \mathcal{X}_T$ is the data instance and $y_{T_i} \in \mathcal{Y}_T$ is the corresponding label. Assume $\mathcal{X}_T'$ and $\mathcal{Y}_T'$ are small portions of data from the target domain, $\mathcal{X}_S'$ and $\mathcal{Y}_S'$ are small portions of data from the source domain. The TL model employed in this study needs $\mathcal{X}_T'$ and $\mathcal{Y}_T'$ to be added to the model training process to fine-tune $F_4(\cdot)$ to acquire $F_5(\cdot)$. The TL model is able to reduce the weights of source instances that are dissimilar to the target instances and increase the weights of source instances that are similar to the target instances (Dai et al., 2007; X. Ma et al., 2024). In this case, $F_5(\cdot)$ could have satisfactory performance for the target domain.

$$F_5([\mathcal{X}_S, \mathcal{X}_T']) = [\mathcal{Y}_S, \mathcal{Y}_T'] \quad (5)$$

Since $\mathcal{X}_T'$ and $\mathcal{Y}_T'$ are not available in this study, $\mathcal{X}_S'$ and $\mathcal{Y}_S'$ are extracted from the source domain to substitute for $\mathcal{X}_T'$ and $\mathcal{Y}_T'$ to form $F_6(\cdot)$. Detailed explanation on how to select $\mathcal{X}_S'$ and $\mathcal{Y}_S'$ are illustrated in section 3.4. In the end, $F_6(\cdot)$ is built for TMC estimation in this study.

$$F_6([\mathcal{X}_S, \mathcal{X}_S']) = [\mathcal{Y}_S, \mathcal{Y}_S'] \quad (6)$$

## 3. METHODOLOGY

### 3.1. Research framework

The main idea behind the proposed framework is to estimate TMCs for intersections. The proposed framework for estimating TMCs using TL consists of several key components, as depicted in the Figure 1. These components are designed to systematically process and utilize data from both source and target domains to estimate TMCs accurately.

In the first step, both source and target domain data undergo feature extraction processes to identify relevant traffic parameters. Feature extraction is a critical step as it transforms raw traffic data into structured information that can be utilized in subsequent steps. Let's assume the whole data set of the source and target domain intersection(s) are denoted as $\mathbb{D}_S$ and $\mathbb{D}_T$, respectively; in this case $\mathbb{D}_S = (\mathbb{X}_S, \mathbb{Y}_S)$ and $\mathbb{D}_T = (\mathbb{X}_T, \mathbb{Y}_T)$. Both $\mathbb{D}_S$ and $\mathbb{D}_T$ have two parts: data instances $\mathbb{X}_S$ and $\mathbb{X}_T$ as well as labels $\mathbb{Y}_S$ and $\mathbb{Y}_T$. $\mathbb{X}_S$ and $\mathbb{X}_T$ contain all the variables extracted from the first step, assume the number of variables extracted is $p$, then $\mathbb{X}_S$ and $\mathbb{X}_T$ can be regarded as $n$ by $p$ matrices.

In the second step, feature selection techniques are employed to identify the most significant features contributing to TMC patterns. This step helps in reducing the dimensionality of the data, enhancing model performance, and minimizing computational complexity. Effective feature selection ensures that the TL models focus on the most informative attributes.

With the important variables selected, a similar-intersection matching algorithm is proposed to match each intersection of the target domain to a certain intersection of the source domain. The core of the TL process involves matching intersections in the source domain with those in the target domain based on their similarity. Similar-intersection matching leverages the selected features to find intersections with analogous traffic characteristics. This matching is crucial for ensuring that the knowledge transferred from the source domain is relevant and applicable to the target domain. Let's assume the number of selected variables is $q$. Let $D_T$ represent data from the target domain intersection and $D_S$ represent data from matched source domain intersection. Similarly, $D_S = (\mathcal{X}_S, \mathcal{Y}_S)$ and $D_T = (\mathcal{X}_T, \mathcal{Y}_T)$. Then both $\mathcal{X}_S$ and $\mathcal{X}_T$ become $n$ by $q$ matrices.

Once similar intersections are identified, data from these source domain intersections are substituted for the target domain intersections. This substitution process effectively creates a synthetic dataset for the target domain, enabling the use of existing data to infer traffic patterns where physical sensors are absent. Target domain data substitution is conducted to extract $\mathcal{X}_S'$ and $\mathcal{Y}_S'$ from the source domain intersection to substitute for $\mathcal{X}_T'$ and $\mathcal{Y}_T'$.

Next, a TL model named Two-stage TrAdaBoost.R2 (henceforth referred to as TrA) is deployed to estimate TMCs for all the intersections in the target domain. The final output of the framework is the estimated TMCs. These predictions provide valuable insights for traffic management authorities, aiding in traffic signal optimization, congestion mitigation, and strategic planning.

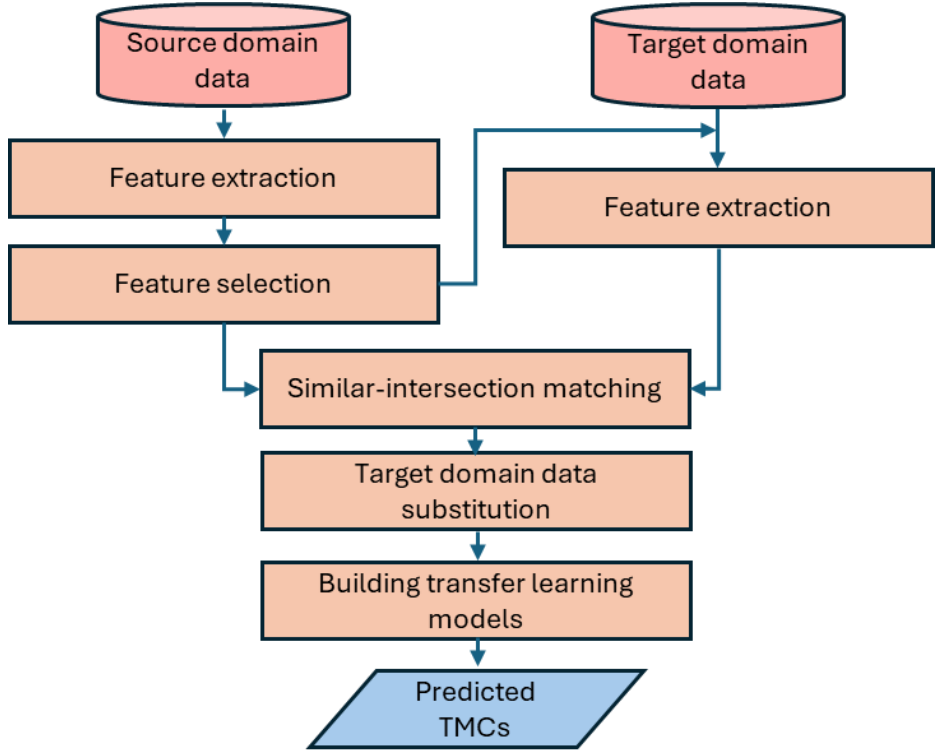

**Figure 1 Research framework**

**3.2. Lasso regression**

In this study, Least Absolute Shrinkage and Selection Operator (Lasso) regression is used for variable importance selection and spatial-temporal association. Lasso regression is a popular technique in the field of statistics and machine learning for both regularization and feature selection in linear regression models. It addresses some of the limitations of traditional linear regression by imposing a penalty on the absolute size of the regression coefficients. This penalty can lead to sparse solutions, where some coefficients are exactly zero, effectively performing variable selection (Tibshirani, 1996).

In a standard linear regression model, we aim to fit a linear relationship between the dependent variable $y$ and a set of independent variables $X = \{x_1, x_2, \ldots, x_p\}$. The model can be represented as:

$$y = \beta_0 + \beta_1 x_1 + \beta_2 x_2 + \cdots + \beta_p x_p + \epsilon \tag{7}$$

Where $y$ is the dependent variable, $\beta_0$ is the intercept, $\beta_i$ (for $i = 1, 2, \ldots, p$) are the regression coefficients, $x_i$ (for $i = 1, 2, \ldots, p$) are the independent variables, $\epsilon$ is the error term.

The goal of linear regression is to find the values of $\beta_0, \beta_1, \ldots, \beta_p$ that minimize the residual sum of squares (RSS):

$$\text{RSS} = \sum_{i=1}^{n}(y_i - \beta_0 - \sum_{j=1}^{p} x_{ij}\beta_j)^2 \tag{8}$$

Lasso regression modifies the objective function of linear regression by adding a regularization term that penalizes the absolute values of the regression coefficients. The objective function for Lasso regression is minimize:

$$\sum_{i=1}^{n}(y_i - \beta_0 - \sum_{j=1}^{p} x_{ij}\beta_j)^2 + \lambda \sum_{j=1}^{p}|\beta_j| \tag{9}$$

where $\lambda$ is a tuning parameter that controls the strength of the penalty. The term $\lambda \sum_{j=1}^{p}|\beta_j|$ is the L1 regularization term, which encourages sparsity in the coefficients $\beta_j$.

Lasso regression is a powerful tool for regularization and feature selection in linear models. By introducing an L1 penalty, it shrinks the regression coefficients and promotes sparsity, making it particularly useful in high-dimensional datasets. Its ability to perform variable selection and reduce overfitting makes it a valuable technique in both statistical analysis and machine learning applications.

### 3.3. Similar-intersection matching

After choosing the important $q$ traffic variables using the lasso regression, the next step is to find a matching function $\mathcal{M}: D_T \rightarrow D_S$ to map each intersection of the target domain to a certain intersection of the source domain. The objective is to find a source intersection that has a similar traffic pattern to the target intersection. To do so, the Pearson Correlation test is calculated pair-wisely between the same traffic variables from the source and target intersections. Finally, for each target intersection, the source intersection with the largest correlation value is selected.

Let's define $\alpha_{S,i}$ and $\alpha_{T,i}$ as the $i$-th traffic variable from source intersections and target intersections. The matched source intersection for each target intersection needs to meet the following criteria:

$$max \sum_{i=1}^{q}(corr(\alpha_{S,i}, \alpha_{T,i})) \tag{10}$$

### 3.4. Target domain data substitution

Recall source domain data $D_S = \{(x_{S_1}, y_{S_1}), \ldots, (x_{S_n}, y_{S_n})\}$, where $x_{S_i} \in \mathcal{X}_S$ is the data instance and $y_{S_i} \in \mathcal{Y}_S$ is the corresponding label. Target domain data is denoted as $D_T = \{(x_{T_1}, y_{T_1}), \ldots, (x_{T_k}, y_{T_k})\}$, where $x_{T_i} \in \mathcal{X}_T$ is the data instance and $y_{T_i} \in \mathcal{Y}_T$ is the corresponding label. When applying TL to induce a predictive model, some labeled data in the target domain are required. However, in practice, labeled data in the target domain are not available. This study chooses an alternative approach that uses labeled source domain data as a substitute for labeled target domain data. The labeled source data used for substitution needs to have the highest similarity to the data instances in the target domain. Specifically, to meet this requirement, $D_S' = \{(x_{S_j}, y_{S_j})\}$ ($D_S' \subsetneq D_S, 1 \leq j \leq m < n$) is created to act as a substitute for the labeled data in the target domain. $D_S'$ only contains pairs $(x_{S_j}, y_{S_j})$ in which the cosine similarity of $x_{S_j}(1 \leq j \leq m < n)$ and $x_{T_i}(1 \leq i \leq k)$ is less than or equal to a threshold value $\vartheta$. $\vartheta$ is set as a hyper-parameter to select the top 10% of data instances that have the highest similarity to $\mathcal{X}_T$ from $\mathcal{X}_S$ (Dai Wenyuan, Y.Q., Guirong, X. and Yong, 2007).

### 3.5. Two-stage TrAdaBoost.R2

As one of the instances-based inductive TL methods, the TrA is a boosting-based algorithm that aims to increase the prediction performance by linearly combining the weak estimators and forming stronger estimators (Pardoe & Stone, 2010; Yehia et al., 2021).

$$F(x) = \sum_{t=1}^{T} \beta_t G(x, \gamma_t) \tag{11}$$

Here $\beta_t$ is the weight of the weak estimator, $G(x, \gamma_t)$ is the weak estimator, and $\gamma_t$ is the optimal parameter of the weak estimator.

Let $D_S$ (of size $n$) denote labeled source domain data and $D'_S$ (of size $m$) denotes labeled data in the target domain. $D$ is the combination of $D_S$ and $D'_S$. $S$ is the number of steps, $F$ is the number of folds for cross-validation. $S$ and $F$ are set as 10 and 5, respectively. The TrA algorithm is shown below.

---

**Input** $D$, $S$, and $F$. Setting the initial weight vector $\mathbf{w}^1$ such that $\omega_i^1 = \frac{1}{n+m}$ for $1 \leq i \leq n + m$

**For** $t = 1, \dots, S$:
1. Call AdaBoost.R2 (Drucker, 1997) with $D$, estimator $G(x)$, and weight vector $\mathbf{w}^t$. $D_S$ is unchanged in this procedure. Obtaining an estimate $error_t$ of $model_t$ using $F$-fold cross-validation.
2. Call estimator $G(x)$ with $D$ and weight vector $\mathbf{w}^t$
3. Calculate the adjusted error $e_i^t$ for each instance as in AdaBoost.R2
4. Update the weight vector

$$\omega_i^{t+1} = \begin{cases} \frac{\omega_i^t \beta_t^{e_i^t}}{Z_t}, & 1 \leq i \leq n \\ \frac{\omega_i^t}{Z_t}, & n+1 \leq i \leq n+m \end{cases}$$

where $Z_t$ is a normalizing constant, and $\beta_t$ is chosen such that the weight of the target (final $m$) is $\frac{m}{n+m} + \frac{t}{(S-1)}\left(1 - \frac{m}{n+m}\right)$

**Output** $F(x) = model_t$, where $t = \mathrm{argmin}_i error_i$

---

## 4. CASE STUDY

### 4.1. Data description

The data utilized in this study was gathered from the eastern Pima County region and encompasses three distinct categories: event-based data, intersection infrastructure data, and Points of Interest (POI) data.

### 4.1.1. Event-based data

The event-based data in this study, collected via the MaxView system—an Advanced Traffic Management System (ATMS)—comprises a series of real-time events. These include detector actuation events, signal change events, pedestrian-related events, and controller communication events. For the purposes of this research, three specific types of event datasets are utilized: detector actuation events, signal change events, and communication events. These datasets are instrumental in deriving event information pertinent to left-turn and through movements, as detailed in Table 1.

### 4.1.2. Infrastructure data

In the proposed method, intersection infrastructure data is manually gathered from Google Maps and encompasses various elements. This data includes the classification of roads (major or minor), the number of left-turn lanes, shared left-turn lanes, through lanes, right-turn lanes, and shared right-turn lanes.

### 4.1.3. POI data

POI data serves as an indicator of urban land use context and economic activities that influence traffic attraction and generation. In this study, the POI data was sourced from the Pima Association of Governments (PAG) employment database. The authenticity of the business presence and the quality of data were validated using the Google Places API and sample reviews (Kramer et al., 2022; Noh et al., 2019). Analysis within this study includes extracting counts of various categories and the number of employees from raw POI data within a 400m radius of an intersection. These category counts provide insights into the diversity of POI and land use characteristics around intersections, while the scale of each POI category is indicated by the employee counts. Together, these metrics are crucial for understanding the traveler characteristics and traffic patterns at nearby intersections.

### 4.2. Experiment design

### 4.2.1. Data sample

The TMC data provided by PAG was an aggregate of 15-minute interval TMC data during peak hours (7:00-9:00 AM, 4:00-6:00 PM) from 2016 to 2020. This data was typically gathered by consulting firms employing a combination of manual data collection methods and individual sensors. On average, data from one to four days per intersection were utilized for model training and validation. Figure 2 presents the layout of the study intersections located in Tucson, Arizona. A total of 30 intersections were selected for inclusion in this study.

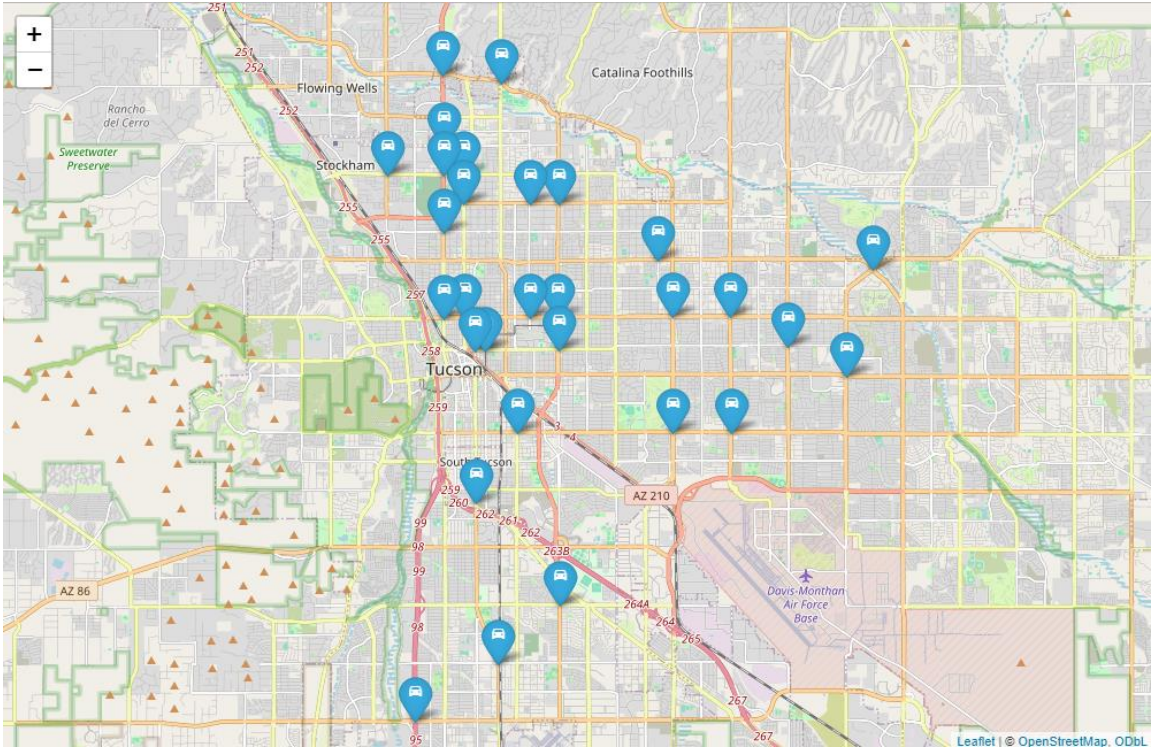

**Figure 2 Layout of the study intersections**

### 4.2.2. Baseline models

Eight state-of-the-art machine learning regression models were chosen as baseline models to assess the feasibility of the proposed framework for estimating TMCs. These models include Support Vector Regression (SVR), K-Nearest Neighbors (KNN), Random Forest (RF), Multi-Layer Perceptron (MLP), Categorical Boosting (CatBoost), Light Gradient Boosting Machine (LightGBM), eXtreme Gradient Boosting (XGBoost), and Adaptive Boosting (AdaBoost). Given the significant impact of hyperparameter settings on the accuracy of each model's estimates, a grid search method was initially developed to identify the optimal hyperparameter values for baseline models and the proposed TL method.

### 4.2.3. Measurements of Effectiveness

Mean Absolute Error (MAE) and Root Mean Square Error (RMSE) are two common criteria used to evaluate and compare prediction methods (Luo et al., 2022). MAE is used to measure the overall errors of the estimation results and RMSE is used to quantify the stability of the estimation results (W. Zhang et al., 2022). These two criteria are employed as performance metrics for comparison in this study and are defined below:

$$\text{MAE} = \frac{1}{N}\sum_{k=1}^{N}|y(k) - \hat{y}(k)| \tag{12}$$

$$\text{RMSE} = \sqrt{\frac{1}{N}\sum_{k=1}^{N}(y(k) - \hat{y}(k))^2} \tag{13}$$

where $y(k)$ is the actual value at time $k$ and $\hat{y}(k)$ is the corresponding predicted value. $N$ is the size of the testing data set (total number of time intervals).

### 4.3. Lasso regression results

Lasso regression ranks the variables that have the highest impact on TMCs. In addition, it provides an approach to better understand the interaction between TMCs, detector data, lane configuration, signal timing, road characteristics, and temporal factors. To better identify the variables, their names are provided.

**Table 2 Lasso Regression Results**

| Variables | Left turn | Through | Right turn |
| --- | --- | --- | --- |
| Through movement detector occupancy time | 0 | 56.82 | 4.20 |
| Through movement detector trigger counts | 0.10 | 18.71 | -2.53 |
| Through movement green time duration | 0 | 8.64 | -2.54 |
| Through movement cycle counts | 2.22 | -4.05 | 0.37 |
| Through movement average of time differences between each pair of consecutive detections | -0.42 | -2.53 | -3.90 |
| Through movement standard deviation of time differences between each pair of consecutive detections | 0 | 6.90 | 0.44 |
| Left-turn movement detector occupancy time | 12.05 | -1.85 | 0 |
| Left-turn movement detector trigger counts | 0 | 1.06 | 1.37 |
| Left-turn movement green time duration | 15.65 | 2.05 | 4.29 |
| Left-turn movement cycle counts | -9.75 | -2.55 | 0 |
| Left-turn movement average of time differences between each pair of consecutive detections | 0 | 0 | 0.85 |
| Left-turn movement standard deviation of time differences between each pair of consecutive detections | 0.20 | 0 | 0.06 |
| Left-turn movement permissive green time | 1.95 | 16.76 | -0.01 |
| Number of shared left turn lanes | 0 | 0 | 0 |
| Number of exclusive left turn lanes | 8.20 | -0.53 | 10.17 |
| Number of through lanes | 4.05 | 44.65 | 1.25 |
| Number of exclusive right turn lanes | 0.19 | 8.84 | 3.04 |
| Number of shared right turn lanes | 0 | 0 | 0 |
| Number of employees of all POI | -0.75 | 21.74 | 1.10 |
| POI categories count | -1.38 | -10.10 | -0.95 |
| Road type | 0 | -41.41 | -3.56 |
| Left-turn type | 0 | 1.59 | -0.32 |
| Minute-of-hour | 0.30 | 1.56 | 0.65 |
| Hour-of-day | -0.47 | 3.59 | 1.39 |

Based on Table 2, it can be seen that through movement detector occupancy time and the number of through lanes are highly influential for through movement counts. The occupancy time has a strong positive impact, indicating that higher detector occupancy correlates with increased through movement counts. Similarly, the number of through lanes also has a positive influence, suggesting that intersections with more through lanes tend to experience higher through traffic volumes. Additionally, the positive coefficients for through movement green time duration and through movement detector trigger counts further emphasize their importance in estimating through movement traffic.

Left-turn movement green time duration and the number of exclusive left turn lanes significantly impact left turn counts. The green time duration for left turns shows a strong positive relationship, meaning longer green times lead to higher left turn counts. The number of exclusive left turn lanes also positively affects left turn counts, indicating that intersections with dedicated left turn lanes experience more left turn traffic. Other variables, such as left-turn movement detector occupancy time, also play a role.

Through movement detector occupancy time, left-turn movement green time duration, and the number of exclusive right-turn lanes notably affect right-turn counts. Higher through movement detector occupancy time positively correlates with right turn counts, likely due to increased overall traffic at the intersection. The green time duration for left turns also positively influences right turn counts, which could be due to coordinated signal timings that benefit multiple movements. Furthermore, having more exclusive right-turn lanes positively impacts right-turn counts, highlighting the importance of dedicated infrastructure for right-turns.

Road type has a strong negative impact on through and right turn counts, indicating that certain road types may be less conducive to high traffic volumes for these movements. This could be due to factors such as road design, traffic regulations, or surrounding land use. Additionally, time-related variables like minute-of-hour and hour-of-day have varying impacts across different movements. For instance, specific times of day may see increased traffic for certain movements due to daily commuting patterns or other temporal factors.

The number of employees of all POI and the number of POI categories have distinct impacts on the TMC estimations for different movements. The number of employees tends to increase through and right turn movements while slightly decreasing left turn movements. The number of POI categories tends to decrease TMC estimations across all movements, with a significant impact on through movements.

Overall, the analysis underscores the complexity of traffic flow at intersections and the necessity of considering multiple variables to accurately estimate traffic counts. The findings highlight the importance of detector data, lane configuration, signal timing, road characteristics, and temporal factors in understanding and managing traffic at intersections.

**4.4. TMCs estimation results**

The inputs for both the baseline models and the proposed TL model were the variables selected by the lasso regression, and the outputs were TMCs with 15-minute intervals. In the model training process, each model was tested using Lasso regression-selected variables, with one intersection being tested at a time and the remaining 29 used for training. This process was repeated 30 times to ensure all intersections were tested. The models evaluated include KNN, SVR, RF, MLP, AdaBoost, XGBoost, CatBoost, LightBoost, and proposed TL. Two measures of effectiveness, MAE and RMSE were used for performance evaluation.

Table 3 shows the estimation performance comparison between eight state-of-the-art regression models and the proposed TL model in terms of MAE. They are the average

MAE values across 30 intersections for each movement type (Left-turn, Through, Right-turn) for all models. The proposed TL model achieved the lowest MAE values of 9.57 for left-turn, 28.59 for through, and 10.25 for right-turn movements, indicating its superior performance in estimating TMCs at intersections. This highlights the significant improvement in estimation accuracy achieved by incorporating TL compared to baseline models such as KNN, SVR, RF, MLP, AdaBoost, XGBoost, CatBoost, and LightBoost.

**Table 3 MAE Comparison for Estimating TMCs of Different Movements**

| Model | Left-turn | Through | Right-turn |
| --- | --- | --- | --- |
| KNN | 19.23 | 76.63 | 20.68 |
| SVR | 16.41 | 80.11 | 18.90 |
| RF | 12.49 | 40.83 | 14.99 |
| MLP | 18.26 | 58.62 | 19.70 |
| AdaBoost | 13.77 | 46.39 | 14.75 |
| XGBoost | 13.12 | 42.60 | 16.87 |
| CatBoost | 12.87 | 41.41 | 15.11 |
| LightBoost | 12.64 | 39.80 | 14.65 |
| TL | **9.57** | **28.59** | **10.25** |

Table 4 compares the RMSE values for different models estimating TMCs at intersections. The average RMSE values across 30 intersections for Left-turn, Through, and Right-turn movements indicate that the TL model demonstrates superior performance across all movement types. Specifically, the TL model achieved the lowest RMSE values of 13.65 for left-turn, 40.62 for through, and 13.98 for right-turn movements. These results highlight the significant improvement in estimation accuracy offered by the TL model over baseline models. The TL model's consistently lower RMSE values underscore its effectiveness in accurately estimating TMCs at intersections.

**Table 4 RMSE Comparison for Estimating TMCs of Different Movements**

| Model | Left-turn | Through | Right-turn |
| --- | --- | --- | --- |
| KNN | 23.59 | 89.46 | 24.51 |
| SVR | 20.47 | 93.64 | 22.68 |
| RF | 16.45 | 51.35 | 18.20 |
| MLP | 22.62 | 70.42 | 23.62 |
| AdaBoost | 17.51 | 56.99 | 18.05 |
| XGBoost | 17.23 | 53.50 | 20.61 |
| CatBoost | 16.60 | 51.34 | 18.34 |
| LightBoost | 16.53 | 49.93 | 17.99 |
| TL | **13.65** | **40.62** | **13.98** |

Figure 3 displays the MAE values for different models estimating TMCs across three movement types: Left-turn, Through, and Right-turn. For left-turn movements, the TL model consistently shows the lowest MAE values, indicating its superior performance.

In contrast, models such as KNN and MLP display higher variability and larger error ranges. For through movements, the TL model again achieves the lowest and most consistent MAE values. Other models, particularly KNN, SVR, and MLP, exhibit higher errors and greater variability, suggesting less reliability in their predictions. For right-turn movements, the TL model maintains its lead with the lowest MAE values, while other models like KNN and MLP show higher error margins. Overall, the TL model demonstrates the most reliable and accurate performance across all movement types, with consistently lower MAE values and reduced variability compared to the baseline models. This visual representation reinforces the effectiveness of the proposed TL in accurately estimating TMCs at intersections.

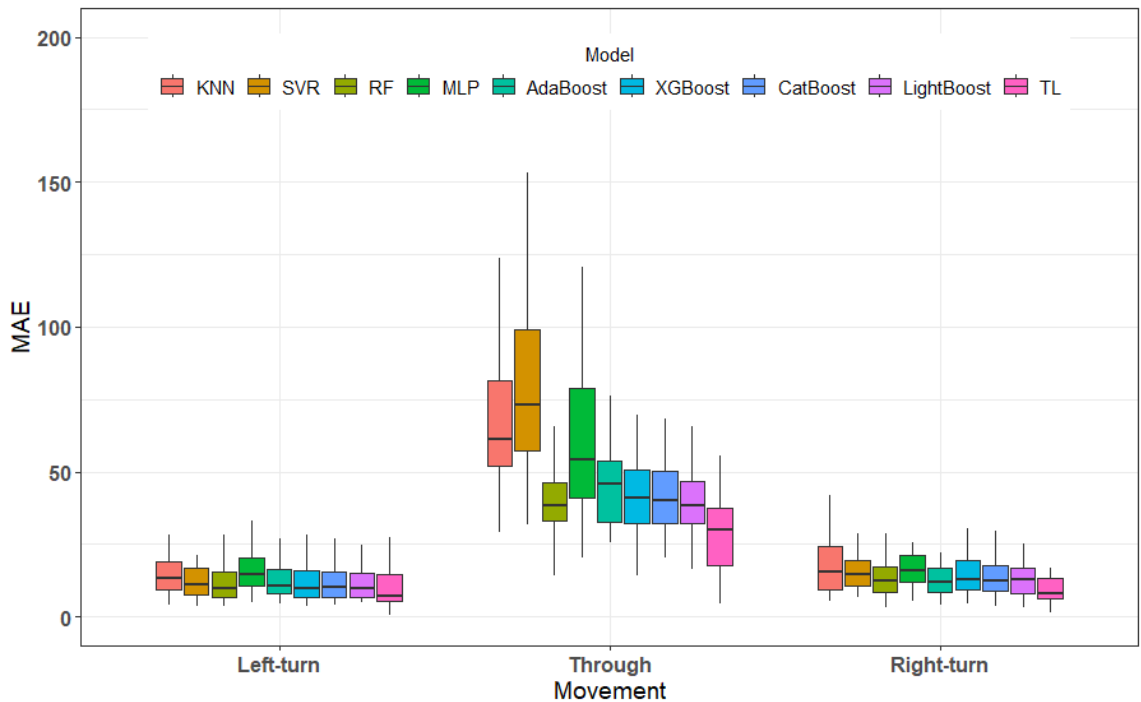

**Figure 3 MAE comparison between different models for different movements**

Figure 4 shows the RMSE values for different models estimating TMCs across three movement types: Left-turn, Through, and Right-turn. For left-turn movements, the TL model exhibits the lowest and most consistent RMSE values, indicating its superior performance compared to other models. Models such as KNN, SVR, and MLP show higher RMSE values, demonstrating less accuracy and higher variability. For through movements, the TL model again outperforms other models with the lowest RMSE values. The baseline models, particularly KNN, SVR, MLP, and AdaBoost, show higher errors, indicating less reliability in their estimations. For right-turn movements, the TL model continues to lead with the lowest RMSE values, while models like KNN and MLP exhibit higher error margins and variability. Overall, the TL model consistently demonstrates the most reliable and accurate performance across all movement types, with lower RMSE values and reduced variability compared to the baseline models. This visual representation underscores the effectiveness of TL in accurately estimating TMCs at intersections without physical sensors, reaffirming its advantage over baseline models.

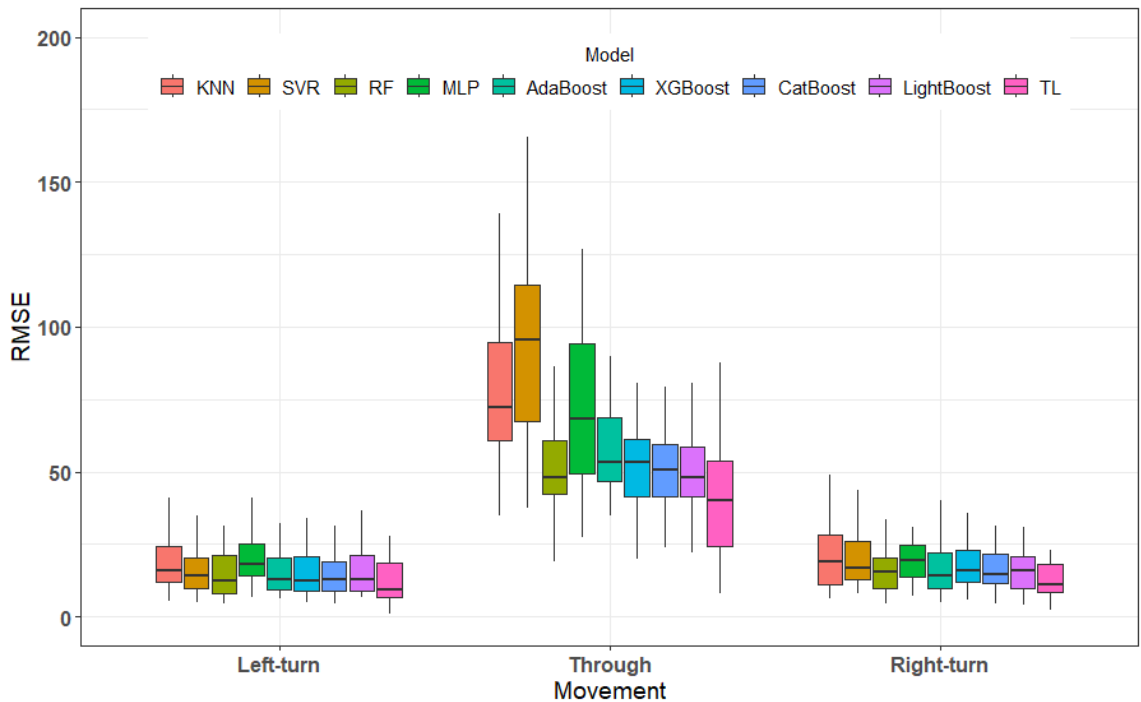

**Figure 4 RMSE comparison between different models for different movements**

Compared to other conventional machine learning models, TL achieves the highest accuracy in estimating TMCs. This superiority is intuitive, as TL relaxes the assumption that the underlying data distributions of the source and target domains must be identical. Additionally, TL effectively handles the estimation of TMCs across various traffic patterns, distributions, and characteristics. Consequently, the proposed TL model can develop scene-specific models, offering superior estimation accuracy compared to generic models.

## 5. CONCLUSION

This research proposes a TL framework for TMC estimation, leveraging data from well-instrumented intersections to infer traffic patterns at intersections lacking sensors. Using traffic controller event-based data, road infrastructure data, and point-of-interest data, the TL framework adapts to diverse traffic patterns, ensuring high accuracy and scalability. Evaluated on 30 intersections in Tucson, Arizona, the proposed method stood out by achieving the lowest MAE values of 9.57 for left-turn, 28.59 for through, and 10.25 for right-turn movements and the lowest RMSE values of 13.65 for left-turn, 40.62 for through, and 13.98 for right-turn movements, indicating its effectiveness in estimating TMCs at intersections.

As the framework has high flexibility allowing for the incorporation of multiple variables, future work could consider adding more temporal and spatial variables to the proposed model. Also, the weather conditions, events or incidents, and other factors that might impact the traffic operation can be included in the model to further improve the generalization ability of the model. This paper is the first attempt to employ TL for TMC estimation. In the future, more advanced machine learning methods and more sophisticated input features could be applied to further improve the estimation performance.